\documentclass[runningheads]{llncs}

\usepackage{placeins}
\usepackage{float}
\usepackage[T1]{fontenc}
\usepackage{graphicx}
\usepackage{amsmath,amssymb}
\usepackage{booktabs}
\usepackage{multirow}
\usepackage{algorithm}
\usepackage{algorithmic}
\usepackage{xcolor}
\usepackage{hyperref}
\usepackage{tikz}
\usetikzlibrary{positioning,arrows.meta,fit,backgrounds,calc}

\begin{document}

\title{StratFormer: Adaptive Opponent Modeling and Exploitation in Imperfect-Information Games}
\titlerunning{StratFormer: Adaptive Opponent Modeling and Exploitation}

\author{
Andy Caen\orcidID{0009-0009-7799-3913} \and
Mark H.M. Winands\orcidID{0000-0002-0125-0824} \and
Dennis J.N.J. Soemers\orcidID{0000-0003-3241-8957}
}

\authorrunning{A. Caen et al.}

\institute{
Department of Advanced Computing Sciences, Maastricht University,\\
Maastricht, The Netherlands\\
\email{a.caen@student.maastrichtuniversity.nl}\\
\email{\{m.winands, dennis.soemers\}@maastrichtuniversity.nl}
}

\maketitle

\begin{abstract}
We present \textsc{StratFormer}, a transformer-based meta-agent that learns to
simultaneously model and exploit opponents in imperfect-information games
through a two-phase curriculum. The first phase trains an opponent modeling
head to identify behavioral patterns from action histories while the agent
plays a game-theoretic optimal~(GTO) policy. The second phase progressively
shifts the policy toward best-response~(BR) exploitation, guided by a
per-opponent regularization schedule tied to exploitability. Our architecture
introduces \emph{dual-turn tokens}---feature vectors constructed at both
agent and opponent decision points---coupled with \emph{bucket-rate features}
that encode opponent tendencies across five strategic contexts. On Leduc
Hold'em, a small poker variant with six cards and two betting rounds, we
test against six opponent archetypes at two strength levels each, with
exploitability ranging from 0.15 to~1.26 Big Blinds (BB) per hand.
\textsc{StratFormer} achieves an average exploitation gain of
$+0.106$~BB per hand over GTO, with peak gains of $+0.821$ against highly
exploitable opponents, while maintaining near-equilibrium safety.

\end{abstract}

\section{Introduction}
\label{sec:intro}

A fundamental tension in imperfect-information games lies between safety and
exploitation. Game-theoretic optimal~(GTO) strategies, grounded in Nash
equilibrium~\cite{zinkevich2007regret}, guarantee worst-case performance but
leave value on the table against suboptimal opponents.
Best-response~(BR) strategies maximally exploit a known opponent but are
themselves highly exploitable if the opponent model is
wrong~\cite{ganzfried2015safe}. Human experts navigate this tradeoff by
gradually modeling opponents during play and adjusting their strategy in real
time~\cite{southey2005bayes}---a capability that has proven difficult to
replicate in learned agents.

Prior work has approached opponent modeling through search-based
methods~\cite{iida1993omsearch}, explicit Bayesian
inference~\cite{southey2005bayes}, policy reconstruction
networks~\cite{he2016opponent}, or theory-of-mind
architectures~\cite{rabinowitz2018machine}. However, these approaches
typically treat opponent modeling and strategic adaptation as separate
modules, creating a gap between understanding the opponent and acting on
that understanding. Reinforcement learning methods such as
LOLA~\cite{foerster2018lola} and neural fictitious
self-play~\cite{heinrich2016deep} learn adaptive strategies but do not
provide interpretable opponent representations.

Recent advances in sequence modeling have shown that transformer architectures
can capture long-range dependencies through self-attention
mechanisms~\cite{vaswani2017attention}. Unlike fixed-state or hand-engineered
opponent models, attention allows the model to dynamically weight past
observations when forming predictions. In imperfect-information games, where
opponent behavior must be inferred from partial and temporally distributed
signals, this ability to selectively attend to relevant past actions provides
a natural foundation for adaptive opponent modeling and strategic adjustment.

We propose \textsc{StratFormer}, a transformer-based meta-agent that unifies
opponent modeling and exploitation in a single architecture through a
two-phase curriculum. In the first \emph{modeling phase}, the agent plays a
GTO policy while training an opponent modeling head to predict opponent
actions from behavioral features. In the second \emph{exploitation phase},
the policy head learns to deviate from GTO toward BR play, with the degree
of deviation controlled by opponent exploitability: highly exploitable
opponents receive aggressive exploitation while near-GTO opponents retain a
higher safety tether. The architecture processes \emph{dual-turn tokens} at
every decision point in the game---both agent and opponent turns---with
causal masking ensuring each token only attends to past information. Five
\emph{bucket-rate features} encode running statistics of opponent behavior
across strategic contexts, enabling the transformer to condition its policy
on a causally-correct opponent model.

While our evaluation focuses on Leduc Hold'em as a tractable testbed where
exact best-responses and exploitabilities can be computed, the architectural
and training contributions are domain-general: dual-turn tokens, the
two-phase curriculum, and the exploitability-tied regularization
$\lambda(\varepsilon)$ apply to any sequential imperfect-information game
with observable opponent actions and a tractable equilibrium baseline. We
evaluate on Leduc Hold'em~\cite{southey2005bayes} against six opponent
archetypes at two strength levels each, spanning a wide range of
exploitability. \textsc{StratFormer} achieves positive exploitation gains
against the majority of opponents while maintaining near-equilibrium safety
against GTO play. We present ablation studies identifying the key design
decisions that enable this balance.

\section{Background}
\label{sec:background}

Imperfect-information games require players to act without full knowledge of
the underlying game state, typically because some variables (such as an
opponent's private cards in poker) are hidden. In two-player zero-sum
settings, a Nash equilibrium guarantees worst-case performance: no opponent
can obtain positive expected value by deviating. GTO play therefore provides
safety, but it is not tailored to a specific opponent and may leave value
unexploited against systematically suboptimal behavior. In contrast, a
best response maximizes expected payoff against a fixed opponent strategy
but can itself become highly exploitable if the opponent model is incorrect.

To quantify this safety--exploitation tradeoff, we measure exploitability
formally. Let $v_i(\sigma_i, \sigma_{-i})$ denote the expected payoff to
player $i$ under strategy profile $(\sigma_i, \sigma_{-i})$, and let
$v_i^*$ denote the equilibrium value. The exploitability of a strategy
$\sigma_i$ is defined as
\[
\varepsilon(\sigma_i)
=
v_{-i}\bigl(\mathrm{BR}_{-i}(\sigma_i), \sigma_i\bigr)
-
v_{-i}^*,
\]
which measures how much additional payoff the opponent can obtain, relative
to equilibrium, by optimally countering $\sigma_i$.

OpenSpiel~\cite{lanctot2019openspiel} reports exploitability via
\emph{NashConv}, the sum of both players' exploitabilities under the
symmetric profile $(\sigma,\sigma)$. In two-player zero-sum games,
\[
\text{NashConv} = 2\,\varepsilon(\sigma),
\]
so we report per-player exploitability as $\text{NashConv}/2$.

Leduc Hold'em~\cite{southey2005bayes} is a two-player poker game with a
six-card deck consisting of two copies each of $J$, $Q$, and $K$. Each player
antes~1~chip and receives one private card. Two betting rounds follow:
preflop (bet size~2) and postflop after a public card is revealed (bet size~4).
Each round permits up to two bets (an initial bet and one raise), with three
legal actions at each decision point: fold, call (or check), and raise (or
bet). The game contains 936 information sets and admits an analytically
computable Nash equilibrium, making it a standard testbed where exact best
responses and exploitability can be computed.

We compute a tabular Nash equilibrium using counterfactual regret
minimization (CFR)~\cite{zinkevich2007regret} as implemented in
OpenSpiel~\cite{lanctot2019openspiel}, running for $10^6$ iterations until
exploitability falls below $10^{-6}$, which is well below the deviations of
even the closest-to-equilibrium evaluation opponent ($\varepsilon = 0.15$).
The resulting strategy is a lookup table mapping each information state to
a probability distribution over legal actions. This tabular GTO serves three
roles in our pipeline: as a pretraining target, as a baseline agent for
paired-seed evaluation, and as the foundation for constructing opponent
archetypes.

\section{Related Work}
\label{sec:related}

The problem of adapting to opponents in games has a long history. Early work
by Iida et al.~\cite{iida1993omsearch} and Donkers~\cite{Donkers_2003_NosceHostem} introduced and built on opponent-model search,
which integrates predictions of opponent behavior directly into game-tree
search. In the poker domain, Southey et al.~\cite{southey2005bayes}
proposed Bayesian opponent modeling using Dirichlet priors over opponent
strategies, and Bard et al.~\cite{bard2013online} extended this with online
implicit agent modeling. He et al.~\cite{he2016opponent} trained neural
networks to predict opponent actions and used these predictions to guide
MCTS rollouts, while Ganzfried and Sandholm~\cite{ganzfried2015safe}
provided theoretical bounds on how much an agent can safely deviate from
equilibrium when exploiting.

In parallel, equilibrium-finding methods have advanced rapidly.
CFR~\cite{zinkevich2007regret} and its
deep variants~\cite{brown2019superhuman,moravvcik2017deepstack} achieve
superhuman play by converging to Nash equilibrium, but these agents play a
fixed strategy and do not adapt to specific opponents. A common thread
across both lines of work is that opponent modeling and strategic adaptation
are treated as separate problems: one module estimates what the opponent is
doing, and another decides how to respond.

On the meta-learning side, MAML~\cite{finn2017model} trains agents that
adapt at deployment, PSRO~\cite{lanctot2017unified} and
LOLA~\cite{foerster2018lola} explore adaptation in multi-agent settings,
and Decision Transformer~\cite{chen2021decision} cast offline RL as
sequence modeling. ToMnet~\cite{rabinowitz2018machine} learns a theory of
mind but uses it separately from its own policy. Our approach differs in
that adaptation is implicit in the transformer's attention over behavioral
history, requiring no gradient steps or separate inference modules at test
time.

\FloatBarrier
\section{Architecture}
\label{sec:architecture}

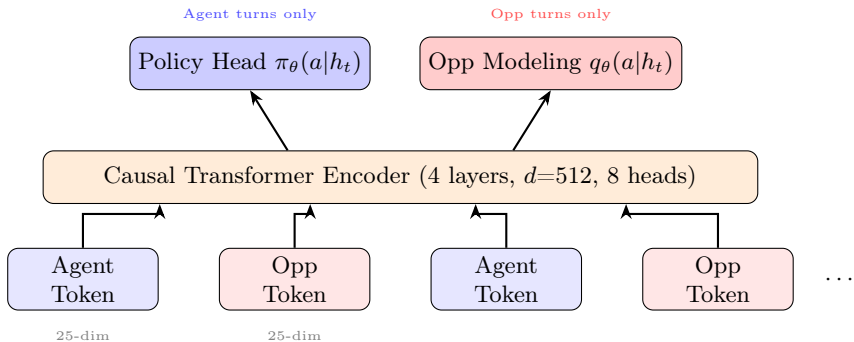
\begin{figure}[ht]
\centering
\begin{tikzpicture}[
    node distance=0.4cm,
    box/.style={rectangle, draw, rounded corners, minimum height=0.7cm,
                minimum width=2.0cm, font=\small, align=center},
    arrow/.style={-{Stealth[length=2mm]}, thick},
    every node/.style={font=\small}
]
\node[box, fill=blue!10] (agent1) {Agent\\Token};
\node[box, fill=red!10, right=0.8cm of agent1] (opp1) {Opp\\Token};
\node[box, fill=blue!10, right=0.8cm of opp1] (agent2) {Agent\\Token};
\node[box, fill=red!10, right=0.8cm of agent2] (opp2) {Opp\\Token};
\node[right=0.3cm of opp2] (dots) {$\cdots$};

\node[below=0.15cm of agent1, font=\tiny, text=gray] {25-dim};
\node[below=0.15cm of opp1, font=\tiny, text=gray] {25-dim};

\node[box, fill=orange!15, minimum width=9.5cm,
      above=1.0cm of $(agent1)!0.5!(opp2)$] (txf)
      {Causal Transformer Encoder (4 layers, $d$=512, 8 heads)};

\node[box, fill=blue!20, above=0.8cm of txf, xshift=-2cm] (policy)
      {Policy Head $\pi_\theta(a|h_t)$};
\node[box, fill=red!20, above=0.8cm of txf, xshift=2cm] (opphead)
      {Opp Modeling $q_\theta(a|h_t)$};

\draw[arrow] (agent1.north) -- ++(0,0.45) -| ([xshift=-3.2cm]txf.south);
\draw[arrow] (opp1.north) -- ++(0,0.45) -| ([xshift=-1.2cm]txf.south);
\draw[arrow] (agent2.north) -- ++(0,0.45) -| ([xshift=1.0cm]txf.south);
\draw[arrow] (opp2.north) -- ++(0,0.45) -| ([xshift=3.0cm]txf.south);

\draw[arrow] ([xshift=-1.5cm]txf.north) -- (policy.south);
\draw[arrow] ([xshift=1.5cm]txf.north) -- (opphead.south);

\node[above=0.1cm of policy, font=\tiny, text=blue!70] {Agent turns only};
\node[above=0.1cm of opphead, font=\tiny, text=red!70] {Opp turns only};
\end{tikzpicture}
\caption{\textsc{StratFormer} architecture. Dual-turn tokens from agent and
opponent decision points are processed by a shared causal transformer. The
policy head outputs the agent's action distribution; the opponent modeling
head is trained via supervised learning to predict the opponent action.}
\label{fig:architecture}
\end{figure}

\textsc{StratFormer} is built on a causal transformer encoder with two
task-specific output heads (Figure~\ref{fig:architecture}). We create a
25-dimensional feature vector at every decision point---both agent and
opponent turns---approximately doubling training tokens per episode and
ensuring causal alignment: each token's features reflect the game state
immediately before the labeled action. Importantly, both agent-turn and
opponent-turn tokens are constructed exclusively from the agent's observable
information (its own cards, the public card, pot size, and the history of
actions). Opponent-turn tokens simply capture this observable state at the
moment the opponent acts, rather than revealing any hidden information. A
binary feature distinguishes the two token types.

\subsection{Feature Engineering}
\label{sec:features}

Each token encodes 25 features organized into three groups. Nine base
features capture card information (agent rank, public rank, normalized),
pot geometry (agent investment, total pot, normalized), action context
(agent's last action, last round action), position flag, round flag, and
observed opponent card (populated after showdown, $-0.5$ sentinel
otherwise). A single turn-type indicator distinguishes agent from opponent
decision points. The remaining fifteen dimensions are running action
frequencies of the opponent across five strategic contexts, with three
values (fold/call/raise rate) per context: global rates across all
decisions, round one and round two rates capturing per-round tendencies,
facing raise rates for situations where the opponent must call or fold to
aggression, and not facing raise rates for voluntary betting positions.
These bucket rates are maintained by a running opponent statistics tracker
that updates incrementally across hands. At test time, the bucket rates
accumulate from the first hand, providing the transformer with increasingly
precise opponent statistics to condition its policy on.

\subsection{Model Architecture and Loss Functions}

The shared encoder~\cite{vaswani2017attention} consists of 4 layers with
$d_{\text{model}} = 512$, 8 attention heads, feedforward dimension~512,
and dropout~0.15. Causal masking restricts each token to attend only to
preceding positions. Input tokens are linearly projected to 512 dimensions,
and sinusoidal positional encodings support sequences up to length~3000
(approximately 375 hands of context).

Two task-specific output heads branch from the shared encoder. The policy
head is a linear projection $\mathbb{R}^{512} \rightarrow \mathbb{R}^{3}$
(fold, call, raise), applied only to agent-turn representations. The
opponent modeling head is a two-layer multi-layer perceptron (MLP)
($512 \rightarrow 512 \rightarrow 3$) with GELU
activation~\cite{hendrycks2016gelu}, applied only to opponent-turn
representations. Separating the heads reduces gradient interference between
the exploitation objective (adjusting the agent's policy) and the
opponent-modeling objective (predicting opponent actions).

The policy head is trained using cross-entropy against a mixture of two
targets: a GTO imitation target and a BR target. A parameter
$\lambda \in [0,1]$ controls the balance between these targets, where
$\lambda = 1$ corresponds to pure GTO imitation and $\lambda = 0$ to pure
best-response. The policy loss is:
\begin{equation}
\mathcal{L}_{\text{policy}} =
\frac{1}{N_h} \sum_{i=1}^{N_h}
\left[
\lambda_i \cdot \mathrm{CE}(\pi_{\text{GTO}}, \pi_\theta)
+
(1 - \lambda_i) \cdot
\mathrm{CE}(\mathbf{1}_{a_{\text{BR}}}, \pi_\theta)
\right],
\end{equation}
where $\mathrm{CE}(p,q) = -\sum_a p(a)\log q(a)$ denotes cross-entropy
with legal-action masking. Both terms use light label smoothing
($\epsilon_s = 0.01$), preventing target probabilities from collapsing to
zero. The GTO term uses cross-entropy rather than Kullback--Leibler (KL)
divergence: cross-entropy only penalizes insufficient coverage of
GTO-supported actions, leaving the model free to allocate additional
probability mass to exploitative moves, while
$\mathrm{KL}(\pi_\theta \,\|\, \pi_\text{GTO})$ would penalize any mass
on non-GTO actions, suppressing the very deviations the BR loss is trying
to learn (we ablate this in Section~\ref{sec:experiments}).

The opponent modeling head is trained via supervised cross-entropy to
predict the opponent's chosen action:
\[
\mathcal{L}_{\text{opp}} =
\frac{1}{N_o}
\sum_{j=1}^{N_o}
\mathrm{CE}(\mathbf{1}_{a_{\text{opp}}^{(j)}}, q_\theta).
\]
The total loss is
\[
\mathcal{L} =
\mathcal{L}_{\text{policy}}
+
\alpha \, \mathcal{L}_{\text{opp}},
\]
where $\alpha$ controls the relative importance of opponent modeling. The
scheduling of $\lambda$ and $\alpha$ during curriculum training is described
in Section~\ref{sec:training}.

The encoder is initialized from a GTO imitation model trained for
15{,}000 epochs via cross-entropy against the tabular equilibrium,
achieving 96\% action agreement. This pretraining stage uses only the
9 base features (without bucket rates), providing a strong equilibrium
baseline so curriculum training focuses on learning controlled deviations.
The input projection is expanded from 9 to 25 dimensions by zero-padding
new columns, preserving pretrained weights while initializing bucket-rate
projections to zero.

\FloatBarrier
\section{Training Methodology}
\label{sec:training}

\subsection{Opponent Archetypes}
\label{sec:archetypes}

We construct opponents by systematically perturbing the tabular GTO
strategy. At each information set where the equilibrium mixes between
actions (i.e., assigns non-zero probability to multiple actions), we
convert the action probabilities to log-odds space, apply a directional
bias, and convert back. Concretely, for a modifier function~$f$ that maps
action indices to bias magnitudes, the perturbed log-odds are
$\ell'_a = \ell_a + w \cdot f(a) + \eta_a$, where $w \in [0, 1]$ controls
deviation strength and $\eta_a \sim \mathcal{N}(0, \sigma^2)$ adds
independent Gaussian noise. Working in log-odds space ensures that the
resulting probabilities remain valid and that small perturbations produce
smooth, realistic deviations rather than degenerate strategies. Information
states where GTO plays a pure strategy (probability~1 on a single action)
are left unmodified, since there is no mixing to perturb.

Six modifier functions define characteristic behavioral directions
(Table~\ref{tab:opponents}). For example, the over-caller modifier
increases the log-odds of the call action and decreases fold, while the
maniac modifier boosts raise at the expense of call and fold. Each
archetype is instantiated at two strength levels (``mid'' $w=0.35$,
``high'' $w=0.70$), selected to produce a broad and roughly linear
distribution of exploitability values across opponent archetypes.
Exploitability values range from $\varepsilon = 0.15$ (passive\_mid, nearly
indistinguishable from GTO) to $\varepsilon = 1.26$ (maniac\_high,
extremely exploitable).

\begin{table}[t]
\centering
\caption{Evaluation opponent suite (mid: $w{=}0.35$, high: $w{=}0.70$).
Exploitability via exact best-response in
OpenSpiel~\cite{lanctot2019openspiel}. All held out from training.}
\label{tab:opponents}
\small
\begin{tabular*}{\linewidth}{@{\extracolsep{\fill}} l l c l @{}}
\toprule
Archetype & Level & $\varepsilon$ & Behavioral bias \\
\midrule
Over-caller & mid / high & 0.23 / 0.53 & Excess calling, never folds \\
Nit & mid / high & 0.34 / 0.48 & Excess folding, plays tight \\
Maniac & mid / high & 0.46 / 1.26 & Excess raising, hyper-aggressive \\
Passive & mid / high & 0.15 / 0.33 & Never raises, passive play \\
Loose aggressive & mid / high & 0.25 / 0.60 & Wide range, frequent raises \\
Tight passive & mid / high & 0.33 / 0.53 & Narrow range, check-calls \\
\bottomrule
\end{tabular*}
\end{table}

\subsection{Two-Phase Curriculum}
\label{sec:curriculum}

Training uses a diverse population of 50 opponents generated with the
log-odds perturbation method. For training, the deviation weight $w$ is
sampled uniformly from the full range $[0,1]$, producing a continuous
spectrum of opponent strengths rather than restricting to the two fixed
mid and high levels used for evaluation. After generating a large
candidate set, we compute exploitability for each opponent, sort them by
$\varepsilon$, and select evenly spaced representatives. This procedure
ensures approximately uniform coverage across the exploitability range
from $\varepsilon = 0.15$ to $1.26$. The evaluation opponents
(Table~\ref{tab:opponents}) are generated with distinct random seeds and
held out entirely from training. Algorithm~\ref{alg:training} summarizes
the two-phase curriculum. Each epoch generates a buffer of 250~hands played
sequentially against the same opponent, so that bucket-rate features
accumulate across hands within the buffer.

\begin{algorithm}[t]
\caption{StratFormer Two-Phase Curriculum Training}
\label{alg:training}
\small
\begin{algorithmic}[1]
\REQUIRE Tabular GTO $\pi_\text{GTO}$, opponent population $\mathcal{O}$,
pretrained model $\theta_0$
\STATE $\theta \leftarrow \theta_0$ \COMMENT{from GTO imitation pretraining}
\STATE \textbf{// Phase 1: Modeling}
\FOR{epoch $= 1$ \TO max\_phase1\_epochs}
    \STATE Sample opponent $o \sim \mathcal{O}$; generate buffer of $N$ hands with agent $= \pi_\text{GTO}$ ($\epsilon$-greedy, $\epsilon{=}0.15$)
    \STATE Record dual-turn tokens with bucket-rate features from opponent tracker
    \STATE Update $\theta$ on $\mathcal{L} = \alpha \cdot \mathcal{L}_\text{opp}$ ($\alpha{=}2.0$)
    \IF{Opp CE $< 0.65$ for 3 consecutive checks}
        \STATE \textbf{break} \COMMENT{modeling converged}
    \ENDIF
\ENDFOR
\STATE \textbf{// Phase 2: Exploitation}
\FOR{epoch $= 1$ \TO remaining\_epochs}
    \STATE Sample opponent $o$ with exploitability $\varepsilon$
    \STATE Compute $\lambda = \lambda_\text{max} \cdot \max(0, 1 - \varepsilon / \varepsilon_\text{max})$
    \STATE Compute BR action $a_\text{BR}$ via tabular best-response to $o$
    \STATE Generate buffer of $N$ hands: agent $= \pi_\theta$; opponent $= o$ (prob 0.9) or $\pi_\text{GTO}$ (prob 0.1)
    \STATE $\mathcal{L}_\text{pol} = \lambda \cdot \text{CE}(\pi_\text{GTO}, \pi_\theta) + (1{-}\lambda) \cdot \text{CE}(\mathbf{1}_{a_\text{BR}}, \pi_\theta)$
    \STATE Update $\theta$ on $\mathcal{L} = \mathcal{L}_\text{pol} + \alpha \cdot \mathcal{L}_\text{opp}$ ($\alpha{=}0.5$)
\ENDFOR
\ENSURE Trained model $\theta$
\end{algorithmic}
\end{algorithm}

The modeling phase trains the opponent modeling head while the agent plays
GTO ($\lambda = 1.0$), skipping best-response computation for approximately
$10\times$ faster data generation. The opponent modeling loss receives
heavy weighting ($\alpha = 2.0$), and the agent plays a uniformly random
legal action with probability $\epsilon = 0.15$ to ensure the modeling head
observes opponent responses to non-equilibrium actions. Ten percent of
buffers use GTO as the opponent. The phase transitions when opponent
modeling cross-entropy drops below~$0.65$ for three consecutive checks, or
after a maximum of 3{,}000~epochs.

Once modeling converges, the agent transitions to exploitation training.
For each opponent with exploitability~$\varepsilon$, the GTO regularization
weight is computed as:
\begin{equation}
\label{eq:lambda}
\lambda(\varepsilon) = \lambda_\text{max} \cdot
\max\!\left(0,\; 1 - \frac{\varepsilon}{\varepsilon_\text{max}}\right)
\end{equation}
with $\lambda_\text{max} = 0.35$ and $\varepsilon_\text{max} = 0.40$. This
gives $\lambda = 0$ (pure best-response) for opponents with
$\varepsilon \geq 0.40$ and a linear taper to $\lambda = 0.35$ for GTO
opponents. Hyperparameters ($\alpha$, $\lambda_\text{max}$,
$\varepsilon_\text{max}$) were chosen via preliminary sweeps on held-out
training opponents to balance exploitation gain and stability.

In Phase~2, the agent always plays the model's own policy during data
generation. Additionally, $10\%$ of training buffers use GTO as the
opponent. Since $\text{BR}(\text{GTO}) = \text{GTO}$, the BR loss for these
buffers provides implicit equilibrium regularization.

Several design choices ensure causal correctness. Training tokens store
independent copies of feature vectors that are never backfilled with the
opponent's card after showdown. A separate inference context (used for
on-policy action selection) receives showdown information normally,
preventing causal leakage into the training data. These safeguards ensure
that the model's adaptation arises strictly from information available at
decision time.

\FloatBarrier
\section{Experiments}
\label{sec:experiments}

We evaluate whether \textsc{StratFormer} successfully balances
opponent-specific exploitation with near-equilibrium safety. Our experiments
measure (i) exploitation gain relative to a tabular GTO baseline and
(ii) performance degradation when facing GTO itself. We further analyze
adaptation dynamics and ablate key design choices.

\subsection{Setup}

We evaluate against the six opponent archetypes described in
Section~\ref{sec:archetypes}, each at two strength levels (12~opponents
total), plus GTO itself; all evaluation opponents are held out from the
training population. We report expected value (EV) per hand in units of
Big Blinds (BB) throughout. To reduce evaluation variance, we use
paired-seed evaluation: a fixed sequence of per-hand random seeds controls
card deals, so both the model and GTO baseline face identical deals and
the measured gain reflects purely strategic differences---analogous to
common random numbers in Monte Carlo simulation~\cite{law2015simulation}.
We train with AdamW~\cite{loshchilov2019decoupled} (learning rate
$3 \times 10^{-5}$, weight decay~0.05, cosine schedule), gradient
accumulation of~8 in Phase~1 and~16 in Phase~2, for 20{,}000~epochs
(1{,}468~gradient updates total) on a single Apple M-series GPU in
approximately 72~hours. Final evaluation uses 3{,}000~hands $\times$
3~independent trials per opponent.

\subsection{Results}

Table~\ref{tab:results} presents held-out evaluation results using
3{,}000~hands $\times$ 3~independent trials per opponent with paired-seed
variance reduction. \textsc{StratFormer} achieves positive exploitation
gain against 8 of 12~opponents, with an average gain of $+0.106$~BB per
hand. Performance against GTO itself averages $-0.050$, indicating that
the model maintains approximate equilibrium safety despite significant
exploitative deviations against weak opponents.

\begin{table}[t]
\centering
\caption{Held-out evaluation (3{,}000 hands $\times$ 3 trials, paired seeds).
Gain $= $ Model~$-$~GTO on identical deals. Bold: 95\% CI excludes zero
($\bar{g} \pm 1.96 \cdot \mathrm{SE}$ across trial means).}
\label{tab:results}
\small
\begin{tabular*}{\linewidth}{@{\extracolsep{\fill}} l r r r r @{}}
\toprule
Opponent ($\varepsilon$) & Model EV & GTO EV & Gain & $\pm$95\% \\
\midrule
GTO (0.00) & $-0.086$ & $-0.037$ & $-0.050$ & --- \\
\midrule
passive\_mid (0.15) & $-0.016$ & $+0.041$ & $-0.058$ & 0.050 \\
over\_caller\_mid (0.23) & $-0.165$ & $+0.006$ & $-0.171$ & 0.025 \\
loose\_agg\_mid (0.25) & $+0.008$ & $+0.028$ & $-0.020$ & 0.050 \\
tight\_passive\_mid (0.33) & $+0.085$ & $+0.014$ & $\mathbf{+0.071}$ & 0.026 \\
passive\_high (0.33) & $+0.117$ & $-0.022$ & $\mathbf{+0.139}$ & 0.011 \\
nit\_mid (0.34) & $-0.050$ & $-0.030$ & $-0.021$ & 0.025 \\
maniac\_mid (0.46) & $+0.131$ & $-0.042$ & $\mathbf{+0.173}$ & 0.031 \\
nit\_high (0.48) & $+0.082$ & $-0.002$ & $\mathbf{+0.084}$ & 0.045 \\
over\_caller\_high (0.53) & $+0.019$ & $+0.006$ & $+0.013$ & 0.102 \\
tight\_passive\_high (0.53) & $-0.000$ & $-0.047$ & $\mathbf{+0.046}$ & 0.038 \\
loose\_agg\_high (0.60) & $+0.220$ & $+0.027$ & $\mathbf{+0.193}$ & 0.030 \\
maniac\_high (1.26) & $+0.839$ & $+0.018$ & $\mathbf{+0.821}$ & 0.074 \\
\midrule
\textbf{Average (excl.\ GTO)} & & & $\mathbf{+0.106}$ & \\
\bottomrule
\end{tabular*}
\end{table}

The strongest exploitation occurs against highly exploitable opponents:
$+0.821$ against maniac\_high ($\varepsilon = 1.26$), $+0.193$ against
loose\_aggressive\_high ($\varepsilon = 0.60$), and $+0.173$ against
maniac\_mid ($\varepsilon = 0.46$). Across opponents, six of twelve gains
are statistically significant at 95\% confidence, with the largest effects
concentrated among the most exploitable archetypes. Four archetypes show
negative or near-zero gains: passive\_mid ($-0.058$, near-GTO with
$\varepsilon = 0.15$ and little value to exploit), loose\_agg\_mid
($-0.020$) and nit\_mid ($-0.021$, both within statistical noise), and
over\_caller\_mid ($-0.171$). The over\_caller\_mid case is the most
informative failure mode: the model loses substantially relative to GTO,
suggesting it over-bluffs against opponents who rarely fold.

The modeling phase transitions at epoch~3{,}499 when opponent modeling
cross-entropy drops below~$0.65$ (from an initial $\ln 3 \approx 1.099$).
After transition, BR cross-entropy decreases from~$0.70$ to~$0.42$,
indicating the policy head is learning exploitative actions, while GTO
cross-entropy rises from~$0.40$ to~$0.79$, reflecting the expected drift
from equilibrium. The opponent modeling cross-entropy continues improving,
stabilizing around~$0.59$ by training end.

The GTO cross-entropy climbs to~$0.79$ by training end, yet the model
achieves approximately $-0.05$ against GTO in held-out evaluation---a
modest loss rather than a collapse, suggesting that near-equilibrium
behavior partially emerges through the best-response pathway
($\text{BR}(\text{GTO}) = \text{GTO}$) rather than explicit imitation alone.

We compare \textsc{StratFormer} against several ablated configurations,
evaluated at the same total training budget (Table~\ref{tab:ablations}).
Without separate phases, the $\lambda{=}0$ variant achieves moderate
exploitation ($+0.111$) but sacrifices GTO safety ($-0.101$); the fixed
$\lambda{=}0.5$ variant retains GTO performance but achieves negative
average gain. Switching the GTO loss to
$\mathrm{KL}(\pi_\theta \,\|\, \pi_\text{GTO})$ eliminates all exploitation
entirely. Reducing $\alpha$ from~$0.5$ to~$0.2$ in Phase~2 causes
exploitation to drop to~$+0.061$ while GTO performance collapses to
$-0.283$, confirming that sustained opponent-modeling gradients are
necessary. Finally, with $\lambda_\text{max} = 0.70$ and
$\varepsilon_\text{max} = 0.70$, the average lambda rises to~$0.33$,
producing near-zero exploitation.

\begin{table}[t]
\centering
\caption{Ablation study. Full model uses held-out evaluation
(Table~\ref{tab:results}); ablated variants use training-time evaluation
(1{,}500 hands, single trial). Relative comparisons remain valid.}
\label{tab:ablations}
\small
\begin{tabular*}{\linewidth}{@{\extracolsep{\fill}} l r r l @{}}
\toprule
Configuration & Avg Gain & GTO EV & Note \\
\midrule
\textsc{StratFormer} (full) & $\mathbf{+0.106}$ & $-0.050$ &
    Two-phase curriculum \\
No phasing ($\lambda{=}0$) & $+0.102$ & $-0.101$ &
    Pure BR, GTO collapses \\
No phasing ($\lambda{=}0.5$) & $-0.042$ & $+0.010$ &
    $\lambda$ too high, no exploit \\
Single-turn tokens & $+0.051$ & $-0.085$ &
    Agent-only tokens \\
KL instead of CE & $-0.004$ & $-0.008$ &
    KL suppresses exploitation \\
Low opp wt.\ ($\alpha{=}0.2$) & $+0.061$ & $-0.283$ &
    Lost modeling, unstable \\
High $\lambda_\text{max}{=}0.70$ & $+0.009$ & $-0.166$ &
    Excessive GTO pull \\
\bottomrule
\end{tabular*}
\end{table}

\FloatBarrier
\section{Conclusion}
\label{sec:conclusion}

We presented \textsc{StratFormer}, a transformer-based meta-agent that
learns adaptive opponent exploitation through a two-phase curriculum. On
Leduc Hold'em, it achieves $+0.106$~BB per hand average gain over a
tabular GTO baseline, with peak gains of $+0.821$ against the most
exploitable opponent. Against GTO itself, the model incurs a $-0.05$~BB
per hand difference, which is small relative to variance at this scale
and does not indicate systematic collapse. Overall, the agent demonstrates
that substantial opponent-specific gains can be achieved while retaining
near-equilibrium behavior.

The main contributors to this performance are:
(i) curriculum phasing that stabilizes opponent modeling before introducing
exploitative gradients,
(ii) dual-turn tokens that align supervision precisely to decision points,
and
(iii) a mixed objective that interpolates between GTO imitation and
opponent-specific best-response targets via $\lambda(\varepsilon)$.

Remaining failure modes highlight the limits of the current objective.
Performance degradation against near-GTO opponents and losses against
call-heavy opponents suggest that accurate opponent modeling alone is
insufficient. The policy must learn how to translate opponent models into
effective counter-strategies, while accounting for how exploitation
incentives change with opponent tendencies and call/fold behavior.

\subsection*{Future Work}

Scaling beyond Leduc requires approximate best-response targets or
search-based targets in larger games such as Texas Hold'em, where exact
tabular BR is infeasible. Richer abstractions and learned value models
will likely be necessary to preserve exploitative precision in
high-dimensional settings.

Two components in the current pipeline are poker-specific: the bucket-rate
features (fold/call/raise frequencies across betting contexts) and the
tabular GTO oracle. Adapting \textsc{StratFormer} to other domains thus
requires re-engineering the per-game behavioral features and substituting
an approximate equilibrium baseline (e.g., Deep CFR~\cite{brown2019deep})
for the tabular Nash. The shared-encoder architecture, dual-turn
tokenization, two-phase curriculum, and $\lambda(\varepsilon)$ schedule
themselves can be reused across environments.

A promising alternative to the current mixed GTO/BR objective is a more
direct reinforcement learning formulation. Rather than interpolating
between two supervised targets---which can produce partially conflicting
gradient signals---future work could optimize expected reward directly
while regularizing by exploitability or distance-to-equilibrium constraints.
Such an approach would allow the agent to discover exploitative deviations
through reward signals, while maintaining safety through explicit penalty
terms or trust-region style bounds.

Additional directions include integrating formal deviation guarantees
(e.g., safe exploitation bounds~\cite{ganzfried2015safe}), adapting to
non-stationary opponents who shift strategies mid-match, and evaluating
against human players rather than synthetic archetypes. Together, these
extensions would test whether the learned opponent representations
generalize beyond controlled experimental settings and support reliable
real-time adaptation in more complex environments.

\clearpage
\bibliographystyle{splncs04}

\end{document}